\pdfoutput=1

\documentclass[11pt]{article}

\usepackage{ACL2023}
\usepackage{times}
\usepackage{latexsym}

\usepackage[T1]{fontenc}

\usepackage[utf8]{inputenc}
\usepackage{color}

\definecolor{dkgreen}{rgb}{0,0.6,0}
\definecolor{gray}{rgb}{0.5,0.5,0.5}
\definecolor{mauve}{rgb}{0.58,0,0.82}
\usepackage{listings}
\usepackage{microtype}

\usepackage{inconsolata}

\usepackage{xcolor}         

\usepackage{tabularx}
\definecolor{bluegray}{rgb}{0.4, 0.6, 0.8}
\definecolor{cornflowerblue}{rgb}{0.39, 0.58, 0.93}
\usepackage{comment} 
\usepackage{subfigure}
\usepackage{graphicx}
\usepackage[ruled,norelsize]{algorithm2e}
\usepackage{algpseudocode}
\usepackage{algorithmicx}
\usepackage[utf8]{inputenc} 
\usepackage[T1]{fontenc}    
\usepackage{hyperref}

\hypersetup{
    colorlinks=true,
    linkcolor=cornflowerblue,
    filecolor=magenta,      
    urlcolor=teal,
    citecolor=teal,
    pdftitle={Benchmarking the Calibration of In-context Learning},
    pdfpagemode=FullScreen,
}

\usepackage{url}            
\usepackage{booktabs}       
\usepackage{amsfonts}       
\usepackage{nicefrac}       
\usepackage{microtype}      
\usepackage{xcolor}         
\usepackage{graphicx}
\usepackage[toc,page,header]{appendix}
\usepackage{minitoc}
\usepackage{makecell}
\usepackage{adjustbox}
\usepackage{colortbl}
\definecolor{Gray}{gray}{0.92}
\usepackage{amsmath}
\usepackage{amssymb}
\usepackage{mathtools}
\usepackage{amsthm}
\usepackage{multirow}

\usepackage{enumitem}
\usepackage{bbm}

\usepackage[capitalize,noabbrev]{cleveref}

\theoremstyle{plain}
\usepackage{mdframed}
\newmdtheoremenv[linewidth=0pt,innerleftmargin=4pt,innerrightmargin=4pt]{prop}{Proposition}

\theoremstyle{definition}

\theoremstyle{remark}

\usepackage{wrapfig}

\definecolor{sclgreyblue}{rgb}{0.2,0.3,0.5}%

\title{Assaying the Robustness of Zero-Shot Machine-Generated Text Detectors}

\author{Yi-Fan Zhang$^{1,2,3}$ \and  Zhang Zhang$^{1,2,3}$ \and  Liang Wang$^{1,2,3}$ \and Tieniu Tan$^{1,2,3}$ \and  Rong Jin$^{4}$ \\
  $^1$University of Chinese Academy of Sciences (UCAS) \\
   $^2$State Key Laboratory of Multimodal Artificial Intelligence Systems (MAIS) \\
  $^3$Institute of Automation, Chinese Academy of Sciences (CASIA)\\
  $^4$ Meta
  \\
  }

\begin{document}
\maketitle
\begin{abstract}
To combat the potential misuse of Natural Language Generation (NLG) technology, a variety of algorithms have been developed for the detection of AI-generated texts. Traditionally, this task is treated as a binary classification problem. Although supervised learning has demonstrated promising results, acquiring labeled data for detection purposes poses real-world challenges and the risk of overfitting. In an effort to address these issues, we delve into the realm of zero-shot machine-generated text detection. Existing zero-shot detectors, typically designed for specific tasks or topics, often assume uniform testing scenarios, limiting their practicality. In our research, we explore various advanced Large Language Models (LLMs) and their specialized variants, contributing to this field in several ways. In empirical studies, we uncover a significant correlation between topics and detection performance. Secondly, we delve into the influence of topic shifts on zero-shot detectors. These investigations shed light on the adaptability and robustness of these detection methods across diverse topics. The code is available at \url{https://github.com/yfzhang114/robustness-detection}.
\vspace{-0.1cm}
\end{abstract}
\vspace{-0.2cm}
\section{Introduction}
Recent strides in natural language generation (NLG) technology have brought substantial enhancements to the diversity, precision, and overall quality of texts generated by LLMs~\cite{zhang2022opt,ouyang2022training,chowdhery2022palm}. Nonetheless, this newfound ability to efficiently create human-like text also brings to the forefront concerns regarding the detection and prevention of LLM misuse in contexts like phishing, disinformation, and academic dishonesty. For instance, numerous educational institutions have imposed bans on ChatGPT due to concerns related to academic integrity, particularly with regard to the potential for students to exploit it for cheating on assignments. Additionally, media outlets have sounded the alarm over the proliferation of fake news generated by LLMs, highlighting the urgent need to address the responsible use of this technology~\cite{floridi2020gpt}. 

In light of these specific usage scenarios, a variety of recent algorithms have surfaced, each tailored for the identification of AI-generated text~\cite{bakhtin2019real,fagni2021tweepfake,kirchenbauer2023watermark}. This detection task is often framed as a binary classification problem~\cite{solaiman2019release,ippolito2019automatic}, aiming to discern text attributes that set apart content created by humans from those generated by language models like LLMs. While supervised learning paradigms have exhibited impressive performance, collecting annotations for detection data can pose real-world challenges, rendering supervised approaches unfeasible in certain situations. Moreover, these approaches present several drawbacks, such as a susceptibility to overfitting to the specific topics it was originally trained on and the requirement to train a fresh model whenever a new topic is introduced.

To address this, we delve into the realm of zero-shot machine-generated text detection, as discussed in~\cite{mitchell2023detectgpt,mireshghallah2023smaller}. In this context, we directly employ the source model without any fine-tuning or adaptation to identify its own generated content. While several existing zero-shot detectors are effective in specific tasks or topics, like question answering or news generation, they often fall short in addressing practical real-world challenges.

Firstly, these detectors typically assume uniform testing instances, where all samples come from the same topic and are equally balanced during evaluation. This assumption proves problematic because, in reality, we lack prior knowledge about the distribution of samples awaiting testing. In extreme cases, where all samples are generated by LLMs, existing zero-shot detectors perform poorly. This underscores the critical importance of acquiring authentic human-authored data beforehand. Although determining a reliable decision threshold through human-authored data is an effective approach, manual data collection is time-consuming and financially impractical for extensive datasets. An alternative strategy involves data extraction from existing human-authored sources, such as websites and scholarly articles~\cite{tang2023science}. Leveraging these readily available sources significantly reduces the time and cost associated with collecting human-authored texts. However, these existing human-authored sources may not precisely match the downstream topics. Therefore, understanding the impact of topic differences on detection performance is crucial for practical applications. Secondly, we must account for the possibility that samples requiring detection may originate from diverse topics, rather than a single uniform topic. For instance, students using ChatGPT to compose essays may choose various subject matters like biographies, science fiction, poetry, and more. Even with reliable human-authored sources, topic shifts are inevitable. In this regard, we make three key contributions to the research community.

1. Across various models, a correlation between topics and detection performance emerges. It becomes evident that documents centered around low-entropy topics generally pose a greater challenge for detection. To deepen our understanding, we delve into the score distributions and conduct a thorough analysis to uncover the underlying reasons. Additionally, it's noteworthy that different methods exhibit considerable variability in their performance across diverse topics. 

2. We investigate the impact of different types of topic shifts on zero-shot detectors. These shifts include transitions from low to high-entropy topics\footnote{Low entropy to high entropy topics refer to instances where documents authored by humans pertain to low-entropy topics and we aim to detect high-entropy documents generated by machines.}, shifts from high to low-entropy topics, combinations of documents from different topics, and more.

3. We explore zero-shot detectors to various state-of-the-art LLMs, such as GPT-2 \cite{radford2019language}, LLaMA \cite{touvron2023llama}, LLaMA-2~\cite{touvron2023llama}, and their variants fine-tuned for instructions or dialogues, such as Alpaca \cite{dubois2023alpacafarm}. This exploration provides valuable insights into the dynamic landscape of AI-generated text and its evolving capabilities.

\vspace{-0.2cm}
\section{Robustness of Zero-Shot Detectors}
\vspace{-0.1cm}

\subsection{Experimental setup}

\textbf{Datasets} For our study, we employ diverse sources of textual data to exemplify various aspects of text detection. Specifically, we use news articles extracted from the XSum dataset~\cite{narayan2018don} as a representation of fake news detection. Wikipedia paragraphs originating from SQuAD contexts~\cite{rajpurkar2016squad} serve as a means to represent machine-generated academic essays. Moreover, we draw on prompted stories sourced from the Reddit WritingPrompts dataset~\cite{fan2018hierarchical} to illustrate the detection of machine-generated creative writing submissions. The python split of CodeSearchNet~\cite{husain2019codesearchnet} is used as machine-generated code detection.

\textbf{Evaluation metrics.} The default metric is the area under the receiver operating characteristic curve (AUROC). Essentially, AUROC represents the probability that a classifier correctly ranks a randomly selected positive example (machine-generated text) higher than a randomly chosen negative example (human-written text). 

\textbf{Methods.} We systematically assess several established zero-shot methods designed for the detection of machine-generated text. These methods rely on the predicted token-wise conditional distributions of the source model to facilitate detection. They are derived from statistical tests based on token log probabilities, token ranks, or predictive entropy, as previously documented in the works of~\cite{gehrmann2019gltr,solaiman2019release,ippolito2019automatic}. Furthermore, our evaluation incorporates the state-of-the-art detection method, DetectGPT~\cite{mitchell2023detectgpt}, which capitalizes on the observation that AI-generated text typically exhibits significantly higher likelihood scores from large language models compared to meaningful perturbations of the same text.

See the appendix for more detailed settings.

\subsection{Measuring topic difference}
To initially explore the connections between topics and detection performance, we suggest employing the concept of average entropy to characterize the categorization of specific subjects. In broad terms, higher entropy indicates a more open-ended nature of generation~\cite{lee2023wrote}.

\textbf{Topic Entropy.} Given a pre-trained language model $\mathcal{P}_\theta(w_t|w_{<t})$, which is trained via maximizing the negative log-likelihood objective $\mathcal{L}=-\mathbb{E}_{t}[\log \mathcal{P}_\theta\left(w_t|w_{<t}\right)]$ on massive corpora. We denote the entropy of a token $w_t$ at position $t$ as $H(w_t|w_{<t}) = - \mathbb{E}_{w_{t}\sim\mathcal{P}_{\theta}(\cdot|w_{<t})}[\log \mathcal{P}_\theta(w_t|w_{<t})]$. We typically measure it based on the answer token $y_n$ (token from the original input sequence) via setting $w_t=y_n$. The topic entropy of dataset $D=\{\mathbf{y}=\{y_j\}_{j=1}^{|y_j|}\}_{i=1}^{|D|}$ is the average entropy across all the sequences.

\subsection{Correlation between detection performances and topics}

\begin{table}[]
\resizebox{0.9\columnwidth}{!}{
\begin{tabular}{ccccc}
\toprule
 & Code & XSum & SQuAD & Writing \\ \hline
Topic Entropy & 2.009 & 2.614 & 2.799 & 3.194 \\
Rank & 0.761 & 0.793 & 0.839 & 0.871 \\
LogP & 0.710 & 0.879 & 0.903 & 0.965 \\
Log Rank & 0.748 & 0.906 & 0.943 & 0.976 \\
DetectGPT &  0.555&  0.767  &0.816  & 0.913  \\ \bottomrule
\end{tabular}%
}
\caption{\textbf{Correlation patterns observed in our data.} The machine-generated texts are created by GPT-2 XL and the reported is AUROC value.}
\label{tab:corr}
\end{table}
We have identified a strong linear correlation between the detection performance of machine-generated content and the topic entropy. The results, as displayed in Table ~\ref{tab:corr}, yield several valuable insights: \textit{\color{sclgreyblue}1. Indicator of Generation Freedom:} Topic entropy effectively serves as an indicator of the degree of creative freedom within each topic. For instance, topics characterized by high entropy, such as stories sourced from the Reddit WritingPrompts dataset, exhibit a greater level of open-endedness compared to low-entropy topics like the Python split of Code-SearchNet. \textit{\color{sclgreyblue}2. Performance Discrepancy:} Zero-shot detectors demonstrate superior performance when applied to higher-entropy topics in contrast to low-entropy topics. This implies that detecting machine-generated content is more challenging in topics where there is lower creative freedom. \textit{\color{sclgreyblue}3. Method Preferences:} It's noteworthy that different detection methods exhibit preferences for specific topics. For example, the Rank method excels in machine-generated code detection but falls short on other high-entropy topics. This suggests that the choice of detection method should be tailored to the characteristics of the topic under consideration. The results considering FPR95 is shown in appendix.

\begin{figure*}[ht]
\begin{minipage}[h]{\textwidth}
\centering
\subfigure[Larger Scale.]{
\includegraphics[width=.3\textwidth]{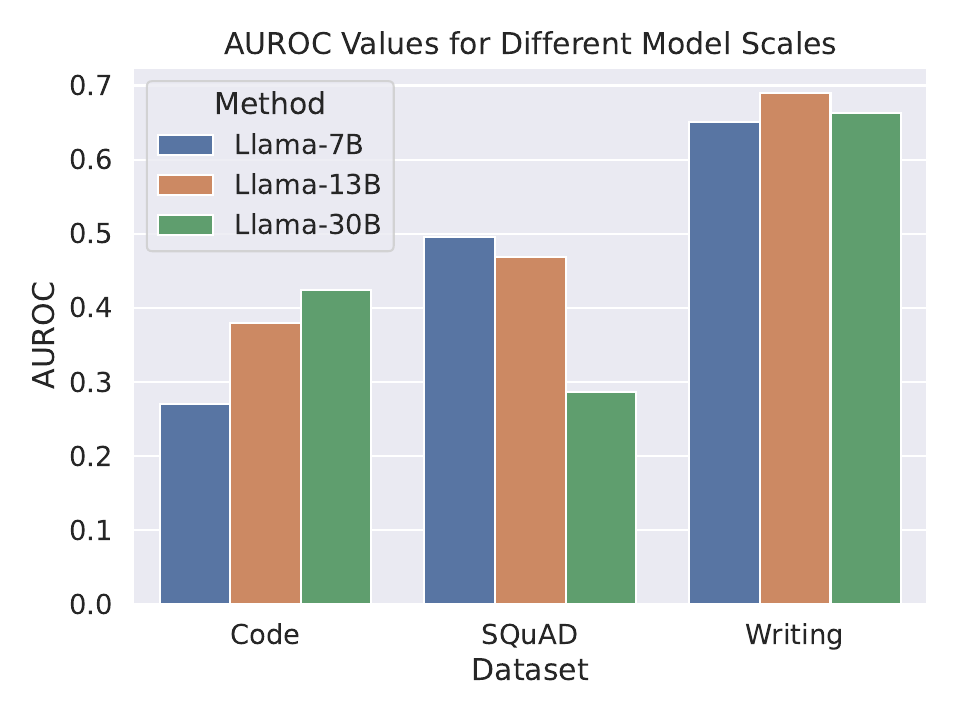}
\label{fig:larger_corr}
}
\subfigure[Topic Transfer.]{
\includegraphics[width=.3\textwidth]{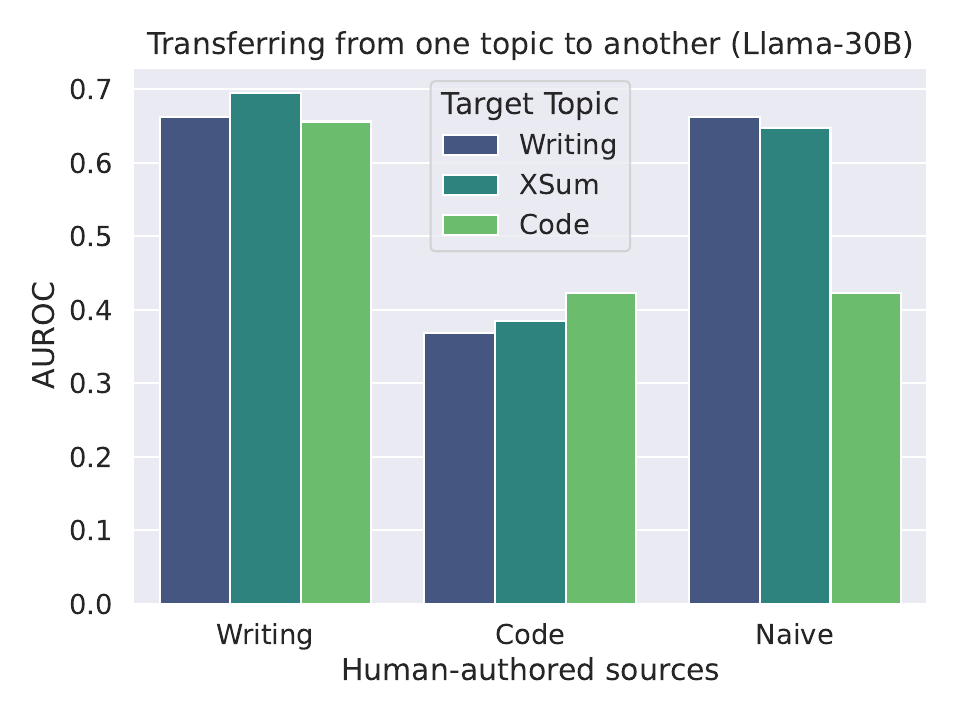} 
\label{fig:larger_transfer} 
}
\subfigure[RLHF-tuned.]{
\includegraphics[width=.3\textwidth]{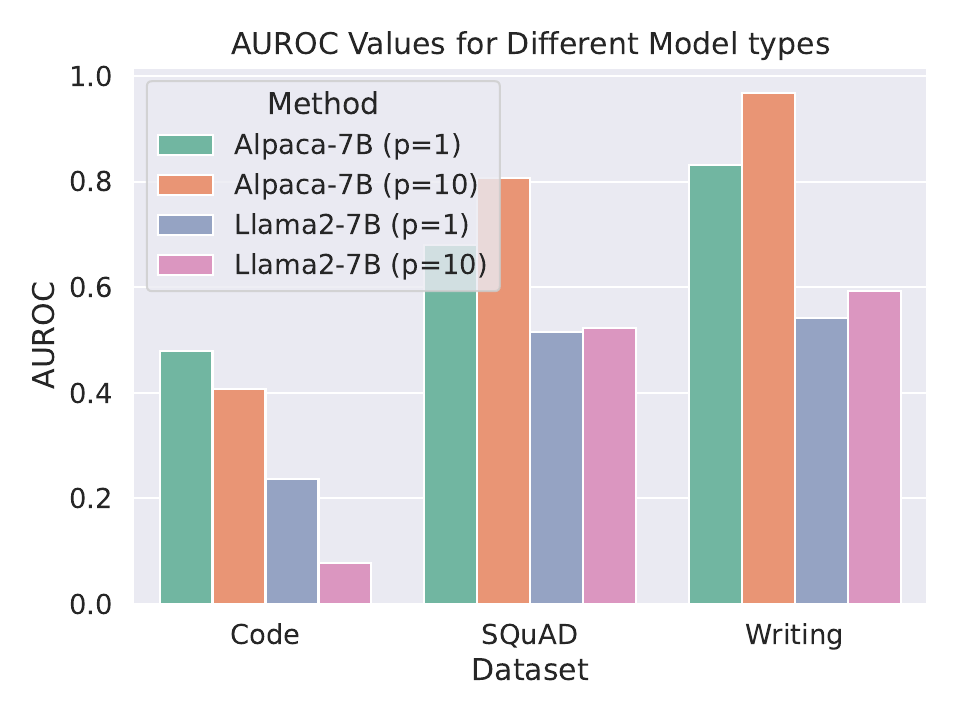} 
\label{fig:rlhf} 
}
\end{minipage}
\caption{
\textbf{Scale up to larger LLMs:} The zero-shot detector is the advanced method DetectGPT and the reported value is AUROC value. $p=i$ means the number of perturbation times of the DetectGPT method.
}
\vspace{-0.2cm}
\label{fig:conf}
\end{figure*}
\subsection{Transfer between different topics}
\begin{table}[]
\centering
\resizebox{0.9\columnwidth}{!}{%
\begin{tabular}{ccccc}
\toprule
Target Topic & Writing & SQuAD & XSum & Code \\ \hline
\multicolumn{5}{c}{\textit{Human-authored sources from Writing}} \\
Rank & 0.871 & 0.915 & 0.924 & 0.861 \\
LogP & 0.965 & 0.984 & 0.992 & 0.983 \\
Log Rank & 0.976 & 0.988 & 0.993 & 0.988 \\
DetectGPT & 0.818  & 0.826 &  0.872& 0.872 \\ \midrule
\multicolumn{5}{c}{\textit{Human-authored sources from Code}} \\
Rank & 0.778 & 0.831 & 0.843 & 0.761 \\
LogP & 0.348 & 0.444 & 0.521 & 0.710 \\
Log Rank & 0.399 & 0.485 & 0.550 & 0.748 \\
DetectGPT & 0.344 & 0.382  & 0.393 &  0.555\\ \bottomrule
\end{tabular}%
}
\caption{\textbf{Detection AUROC when transferring from one topic to another}.}
\label{tab:topic_shift}
\end{table}
\begin{table}[]
\resizebox{\columnwidth}{!}{%
\begin{tabular}{ccccccc}
\toprule
 & \multicolumn{3}{c}{Writing (W) and XSum (X)} & \multicolumn{3}{c}{Writing (W) and Code (C)} \\ \cmidrule{2-4} \cmidrule{5-7}
Method & W+X-W & W+X-X & W+X-W+X & W+C-W & W+C-C & W+C-W+C \\
Rank & 0.794 & 0.858 & 0.826 & 0.825 & 0.811 & 0.818 \\
LogP & 0.822 & 0.935 & 0.879 & 0.656 & 0.847 & 0.751 \\
Log Rank & 0.876 & 0.950 & 0.913 & 0.688 & 0.868 & 0.778 \\ \bottomrule
\end{tabular}%
}
\caption{\textbf{AUROC when transferring among mixtures of topics.} A-B : human-authored texts are from topic A and the machine-generated texts are from B.}
\label{tab:mixture}
\vspace{-0.2cm}
\end{table}
In this subsection, we conduct an exploration of the influence of topic shifts on a variety of zero-shot detectors. We maintain the consistency of human-authored sources while introducing variability in the source of the model-generated text to evaluate each method's performance. Key findings from our analysis, detailed in Table ~\ref{tab:topic_shift}, emphasize several significant insights: \textit{\color{sclgreyblue}{1. Topic Entropy and Detection:}} When presented with high-entropy human-authored text, the task of identifying low-entropy machine-generated text becomes relatively straightforward, yielding superior performance. However, this situation is notably reversed when both data types exhibit high entropy.  \textit{\color{sclgreyblue}2. Sensitivity to Distribution Shifts:} Certain methods, including log p and log rank, which stand out as effective baselines, are remarkably sensitive to distribution shifts. This sensitivity manifests when human-authored text has high entropy, while machine-generated text displays low entropy, causing a substantial decline in their performance. Surprisingly, methods like Rank, which may not exhibit exceptional individual performance, prove to be highly sensitive to topic shifts compared to other techniques. \textit{\color{sclgreyblue}3. Amplified Effects with Topic Entropy Disparity:} As the difference in topic entropy widens, the pronounced nature of these three trends becomes even more apparent. These findings emphasize the critical importance of comprehensive data topic analysis before the deployment of a model.

\textbf{Examining Mixed Topics}. Table ~\ref{tab:mixture} distinctly reveals a trade-off related to topic combinations. To be specific, the detection performance for a blend of Writing and Code documents outperforms that of only Code instances but falls behind when compared to the detection on Writing instances exclusively. Additionally, introducing human-authored documents from low-entropy topics appears to further exacerbate the detection challenges.

\begin{figure}[b]
\begin{minipage}[h]{\linewidth}
\centering
\subfigure[LogP of Code.]{
\includegraphics[width=.45\linewidth]{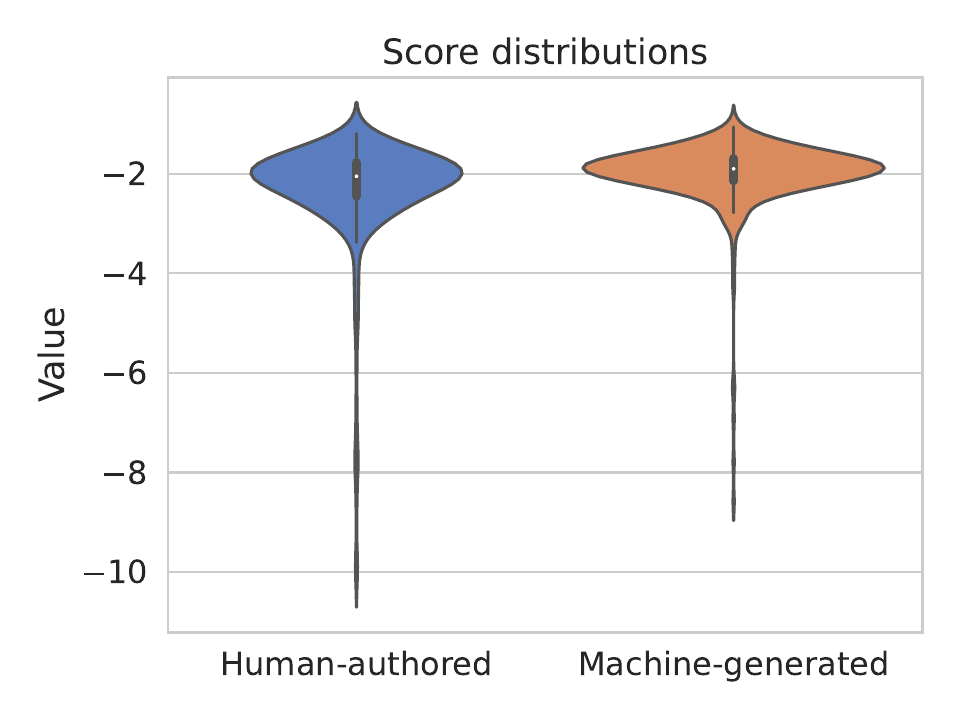}
\label{fig:dis_logp_code}
}
\subfigure[LogP of Writing.]{
\includegraphics[width=.45\linewidth]{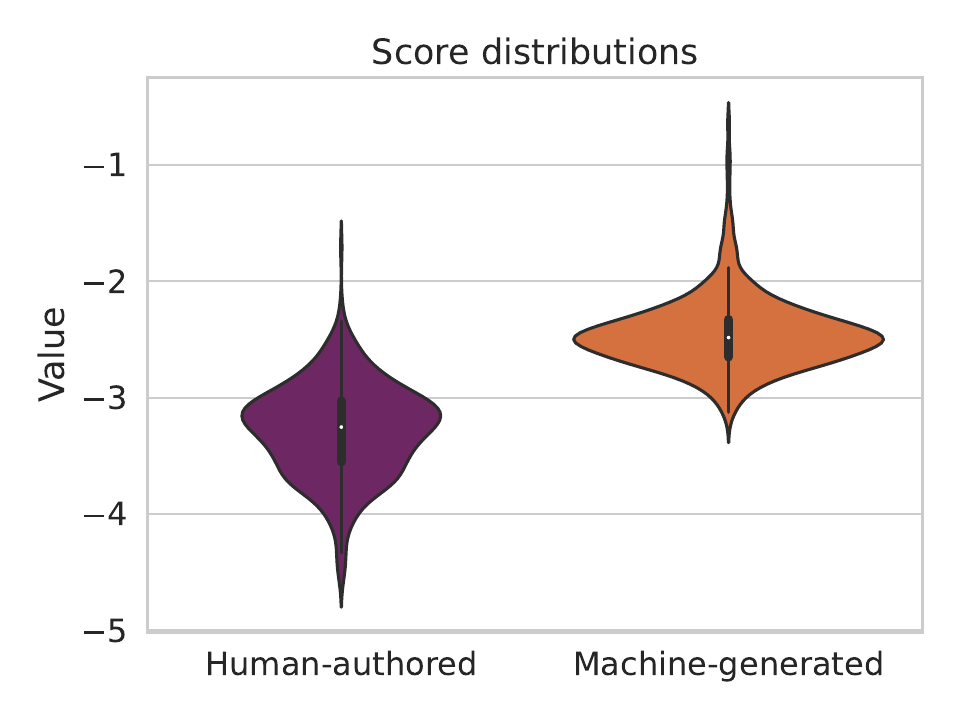} 
\label{fig:dis_logp_writing} 
}
\end{minipage}
\caption{
\textbf{Score distributions} of log probabilities considering different topics.
}
\vspace{-0.2cm}
\label{fig:dis}
\end{figure}

\textbf{Investigate Root Causes}: In Figure \ref{fig:dis}, an examination of the log probability distributions for Code and Writing reveals that the mean values of the two score distributions for Code are challenging to differentiate, resulting in suboptimal performance. Additionally, the variance of scores for low-entropy topics is larger, further complicating the task. This phenomenon elucidates why combining different topics merely diversifies the score distribution without adding value to detection performance. Refer to Figure \ref{fig:app_socres} for a comprehensive visualization.

\subsection{Scale up to larger LLMs}
We delve into an examination of the persistence and evolution of the aforementioned insights across various LLMs. Illustrated in~\figurename~\ref{fig:conf}, we can observe that these insights, such as the correlation between detection performance and topic entropy (\figurename~\ref{fig:larger_corr}), and the phenomenon of topic transfer (\figurename~\ref{fig:larger_transfer}), remain valid even as we scale up LLMs. It's intriguing to note that the relationship between detection performance and model size is not a straightforward linear one. Instead, we identify distinct trends for different topics. Additionally, the Reinforcement Learning from Human Feedback (RLHF) models exhibit varying behaviors in contrast to conventional LLMs. It appears that models fine-tuned through reinforcement learning from human feedback are notably more susceptible to detection~\figurename~\ref{fig:rlhf}. Furthermore, in the context of advanced detection methods, like DetectGPT, subjecting the model to multiple perturbations doesn't consistently yield benefits (Table.~\ref{tab:detectgpt-large}), particularly when applied to low-entropy topics.

\section{Conclusions and Limitations}
Within this paper, we perform a comprehensive assessment of established zero-shot machine-generated text detectors across various types of topic shifts. We establish a strong correlation between detection performance and topic characteristics, revealing that texts with higher degrees of openness are more readily detectable. Subsequently, we explore diverse scenarios of topic shifts and ascertain that these existing detectors are all notably responsive to topic alterations, albeit with varying levels of resilience. In conclusion, we extend our investigation to encompass advanced LLMs, including more extensive models like LlaMA2 and those fine-tuned by RLHF. Our intention is for these empirical findings to provide valuable insights for method selection and the design of studies pertaining to zero-shot detectors. 

\textbf{Limitations.} Due to space limits, the study may not cover the entire spectrum of possible topics, and different models not included in the study may exhibit distinct behaviors, impacting the overall applicability of the results.

\bibliography{acl2023}
\bibliographystyle{acl_natbib}

\newpage
\appendix

\section{Appendix}
\label{sec:appendix}

\subsection{Experimental settings}

\textbf{Dataset selection.} This selection of datasets allows us to explore and evaluate text detection across different domains and applications. Within each experiment, the evaluation encompasses a sample size ranging from 150 to 500 instances, as detailed in the text. Throughout these experiments, the process of generating machine-generated text initiates with the first 30 tokens extracted from the original text. \textit{Instances Authored by Humans and Their Machine-Generated Counterparts} for various topics are presented in Table~\ref{app:datasets}.

\textbf{Evaluation metrics.} In addition to AUROC, we also consider the False-Positive Rate (FPR95), which signifies the rate of false positives among machine-generated samples when $95\%$ of human-generated samples are accurately detected. FPR95 holds particular importance for groups that produce unconventional text, such as non-native speakers, as they may be more susceptible to false positives. This is of significant concern, especially if these detectors are applied in educational settings. It's important to note that, unless otherwise specified, all experiments maintain a balanced distribution of positive and negative examples. 

\textbf{Implementation details.} All of our experiments are based on Pytorch~\cite{paszke2019pytorch}. We conduct all the experiments on a machine with Intel R Xeon (R) Platinum 8163 CPU @ 2.50GHZ, 32G RAM, and Tesla-A100 (32G)x4 instances.

\textbf{License.} All the assets (i.e., datasets and the codes for baselines) we use the MIT license containing a copyright notice and this permission notice shall be included in all copies or substantial portions of the software.

\subsection{Extended results}

\textbf{Correlation Between Detection Performance and Topics.} Examining the FPR95 metric reveals similar observations, as detailed in Table~\ref{tab:corr_fpr}.

\begin{table}[h]
\resizebox{0.9\columnwidth}{!}{
\begin{tabular}{ccccc}
\toprule
 & Code & XSum & SQuAD & Writing \\ \hline
Topic Entropy & 2.009 & 2.614 & 2.799 & 3.194 \\
Rank & 0.638 & 0.580 & 0.413 & 0.402 \\
LogP & 0.786 & 0.600 & 0.523 & 0.262 \\
Log Rank & 0.760 & 0.546 & 0.330 & 0.126 \\
DetectGPT (p=1) & 0.922 & 0.824 & 0.737 & 0.682 \\
DetectGPT (p=10) & 0.485 & 0.356 & 0.273 & 0.164 \\
 \bottomrule
\end{tabular}%
}
\caption{\textbf{Correlation patterns observed in our data.} The machine-generated texts are created by GPT-2 XL and the reported is FPR95 value. $p=i$ means the number of perturbation times of the DetectGPT method.}
\label{tab:corr_fpr}
\end{table}

\begin{table}[]

\begin{tabular}{lccc}
\toprule
\multicolumn{4}{c}{\textit{Human-authored sources from Writing}} \\ \hline
Target Topic & \multicolumn{1}{l}{Writing} & \multicolumn{1}{l}{XSum} & \multicolumn{1}{l}{Code} \\
DetectGPT & 0.663 & 0.695 & 0.656 \\
DetectGPT (p=10) & 0.754 & 0.778 & 0.742 \\ \midrule
\multicolumn{4}{c}{\textit{Human-authored sources from Code}} \\
Target Topic & \multicolumn{1}{l}{Writing} & \multicolumn{1}{l}{XSum} & \multicolumn{1}{l}{Code} \\
DetectGPT & 0.369 & 0.385 & 0.423 \\
DetectGPT (p=10) & 0.294 & 0.324 & 0.392 \\ \bottomrule
\end{tabular}%
\caption{\textbf{Detection AUROC when transferring from one topic to another}. The machine-generated texts are created by \textbf{Llama-30B}
and $p=i$ means the number of perturbation times of the DetectGPT method.}
\label{tab:detectgpt-large}
\end{table}

\begin{table*}[]
\resizebox{\linewidth}{!}{%
\begin{tabular}{lcc}
\bottomrule
\textbf{Dataset} & \textbf{Human-authored} & \textbf{Machine-generated} \\ \hline
Writing &  \multicolumn{1}{|m{10cm}|}{An only son takes his mother on an epic road trip before Alzheimers takes her memory away completely. "Take my picture, Ron." She said holding a pancake at a dive in Iowa, just like she had in front of the grand canyon, with the dancers in Las Vagas, at LAX, and with every meal she had eaten in between. Snap. "Now a picture of yourself." I turned the phone around and took a selfie with my breakfast, just as I had done at the golden gate bridge, the coast of Oregon, Yellowstone National Park, and every other stop that my mother and I had made. I noticed over the past few weeks that a haze had began to claim more and more of her memory, but every fight she had inside of her was arm wrestling.}
 & \multicolumn{1}{|m{10cm}|}{An only son takes his mother on an epic road trip before Alzheimers takes her memory away completely. "Take my picture, Ronnie. You'll need to remember this." Ronnie is a 60-year-old man who has been caring for his mother, Rose, for the last 10 years. She was diagnosed with Alzheimer's at the age of 50, and since then Ronnie has been her sole caretaker. Rose has always been a strong-willed woman, and Ronnie has always been her son. Now, as Rose's condition worsens, Ronnie decides to take her on a road trip to California. Ronnie wants to take his mother on a trip so that she can see the ocean before her memory is completely gone. Ronnie is determined to make this trip happen, but he knows it won't be easy. Ronnie's mother is stubborn and} \\ \hline
SQuAD & \multicolumn{1}{|m{10cm}|}{An adolescent's environment plays a huge role in their identity development. While most adolescent studies are conducted on white, middle class children, studies show that the more privileged upbringing people have, the more successfully they develop their identity. The forming of an adolescent's identity is a crucial time in their life. It has been recently found that demographic patterns suggest that the transition to adulthood is now occurring over a longer span of years than was the case during the middle of the 20th century. Accordingly, youth, a period that spans late adolescence and early adulthood, has become a more prominent stage of the life course. This therefore has caused various factors to become important during this development. So many factors contribute to the developing}
& \multicolumn{1}{|m{10cm}|}{An adolescent's environment plays a huge role in their identity development. While most adolescent studies are conducted on white, middle class, heterosexual males, it is important to understand how identity development differs for those who do not fit this mold. This study examined the identity development of 14 adolescents who identify as lesbian, gay, bisexual, or transgender (LGBT) and 14 adolescents who identify as heterosexual. The sample was recruited through an online survey and was composed of 14 males and 14 females. The results indicated that the LGBT adolescents were more likely to experience a greater sense of alienation from their peers than the heterosexual adolescents. The LGBT adolescents also had a greater sense of alienation from their families. The heterosexual adolescents had a greater} \\ \hline
XSum & \multicolumn{1}{|m{10cm}|}{North, 24, landed on his head after a high tackle from Adam Thompstone in his side's 19-11 defeat against Leicester Tigers on Saturday, 3 December. "I read some reports he must finish and stop playing. He doesn't want to be treated like that," Mallinder said. "All George wants to do is get back and play rugby." The Wales international previously had a six-month spell out of the game after suffering a series of blows to the head during matches, including a serious concussion when scoring a try against Wasps in March 2015. Media playback is not supported on this device The former Scarlets winger will sit out Northampton's game against} & \multicolumn{1}{|m{10cm}|}{North, 24, landed on his head after a high tackle from Adam Thompstone in his side's 19-14 defeat to Leeds Rhinos. He was stretchered off and taken to hospital where he was diagnosed with a fractured skull. The injury has been described as "very serious" by North's club, the Castleford Tigers, and he is likely to miss the rest of the season. A statement on the club's website read: "The club can confirm that it is very serious and that the player is likely to be out for the rest of the season." North, who joined the Tigers from the Wakefield Trinity Wildcats last year, has scored six tries in 13 appearances this season.} \\ \hline
Code & \multicolumn{1}{|m{10cm}|}{Code~\ref{pythoncode1}} & \multicolumn{1}{|m{10cm}|}{Code~\ref{pythoncode2}}\\
 \bottomrule
\end{tabular}%
}
\caption{Machine generation samples for the 4 datasets, and used language models is Llama-7B. }\label{app:datasets}
\end{table*}
\newpage
\begin{lstlisting}[language=Python, frame=single, caption={Human-authored}, label=pythoncode1]
def cli(env, prop):
    """Find details about this machine."""

    try:
        if prop == 'network':
            env.fout(get_network())
            return

        meta_prop = META_MAPPING.get(prop) or prop
        env.fout(SoftLayer.MetadataManager().get(meta_prop))
    except SoftLayer.TransportError:
        raise exceptions.CLIAbort(
            'Cannot connect to the backend service address.')
    \end{lstlisting}
 \begin{lstlisting}[language=Python, frame=single, caption={Machine-generated}, label=pythoncode2]
def cli(env, prop):
    """Find details about this machine."""

    try:
        if prop == 'network':
            return env.get('network', {})
        elif prop == 'storage':
            return env.get('storage', {})
        elif prop == 'cpu':
            return env.get('cpu', {})
        elif prop == 'memory':
            return env.get('memory', {})
        elif prop == 'disk':
            return env.get('disk', {})
    \end{lstlisting}

\begin{figure*}[ht]
\begin{minipage}[h]{\textwidth}
\centering
\subfigure[LogP of Code.]{
\includegraphics[width=.23\textwidth]{figures/metric_datasets/metric_dis_Log_p_code.pdf}
}
\subfigure[LogP of XSum.]{
\includegraphics[width=.23\textwidth]{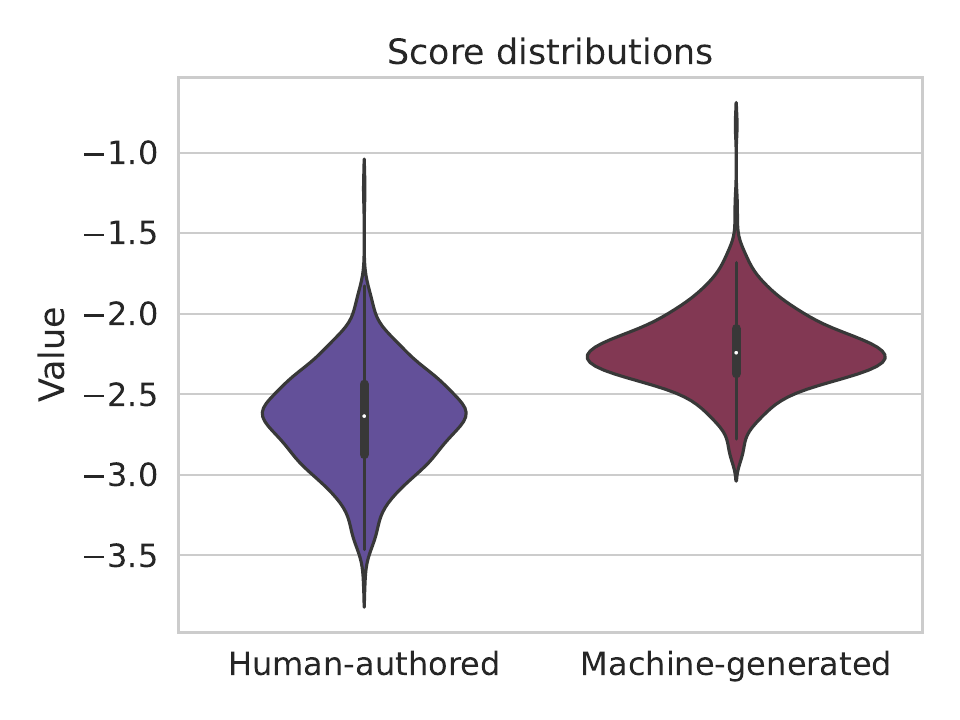} 
}
\subfigure[LogP of SQuAD.]{
\includegraphics[width=.23\textwidth]{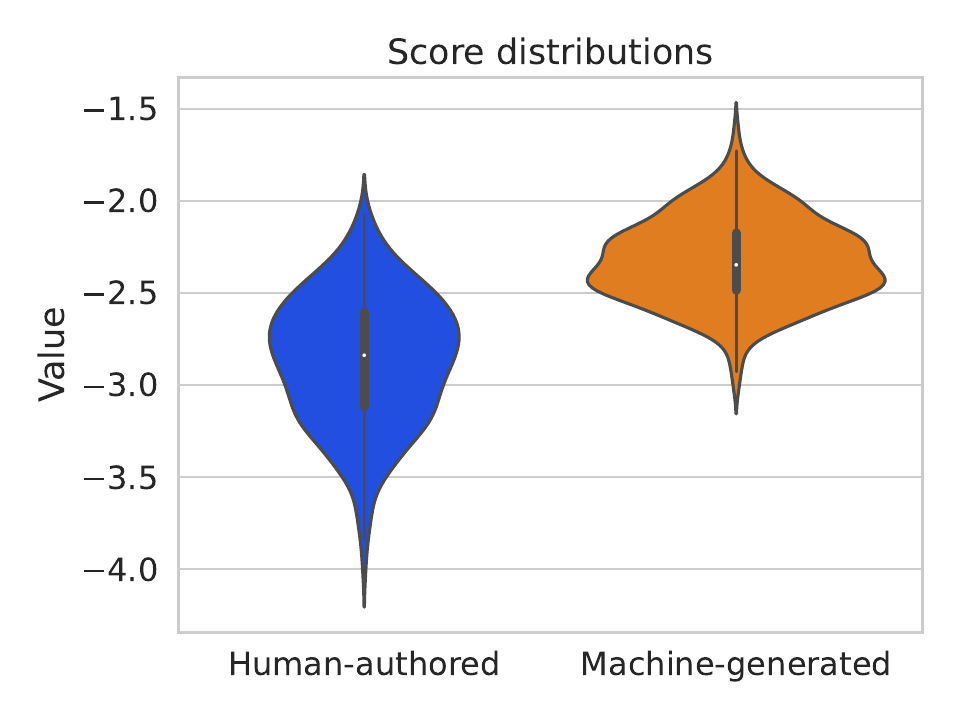} 
}
\subfigure[LogP of Writing.]{
\includegraphics[width=.23\textwidth]{figures/metric_datasets/metric_dis_Log_p_writing.pdf} 
}

\subfigure[Rank of Code.]{
\includegraphics[width=.23\textwidth]{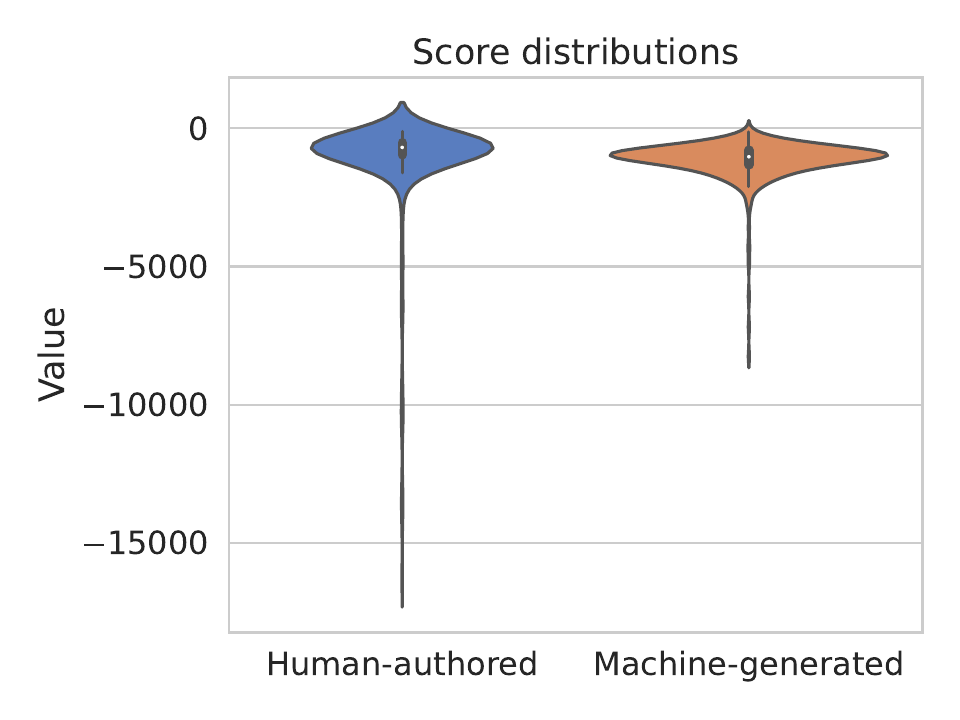}
}
\subfigure[Rank of XSum.]{
\includegraphics[width=.23\textwidth]{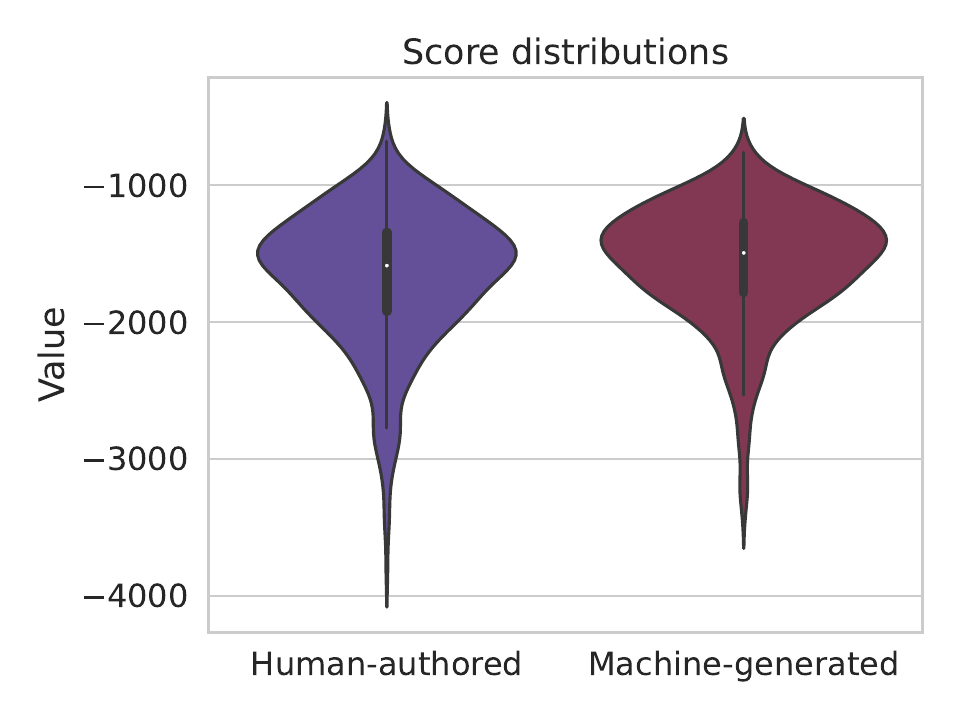} 
}
\subfigure[Rank of SQuAD.]{
\includegraphics[width=.23\textwidth]{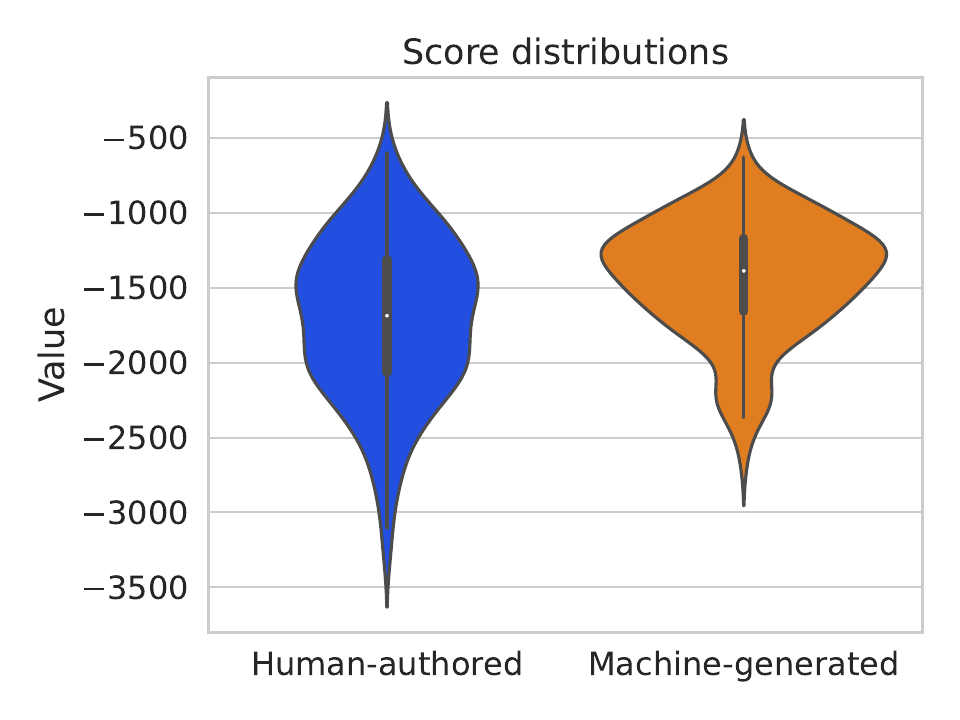} 
}
\subfigure[Rank of Writing.]{
\includegraphics[width=.23\textwidth]{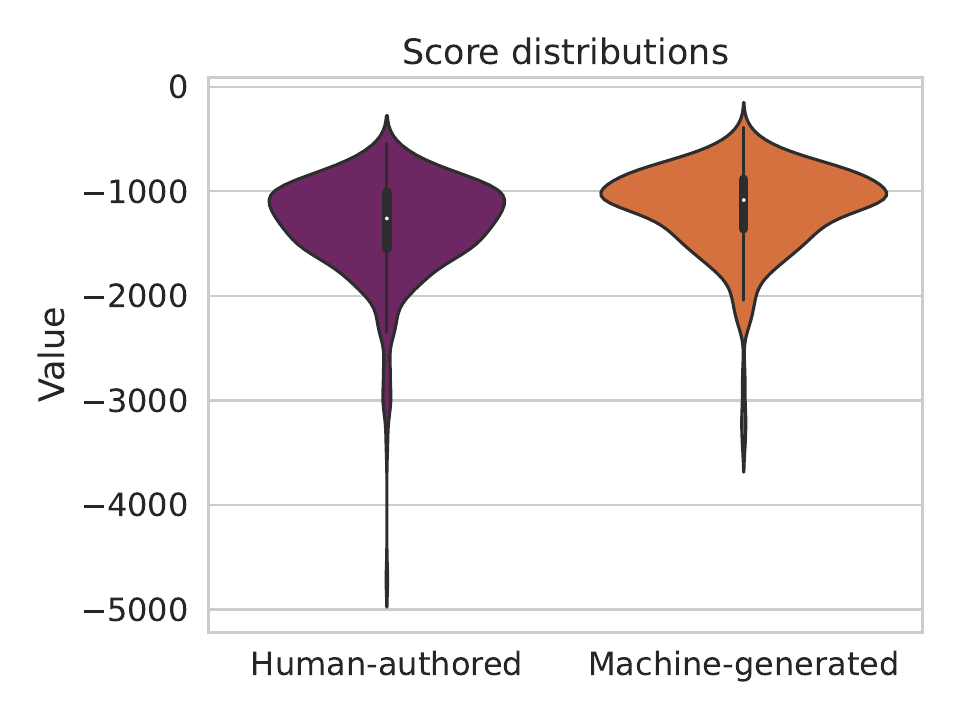} 
}

\subfigure[Log Rank of Code.]{
\includegraphics[width=.23\textwidth]{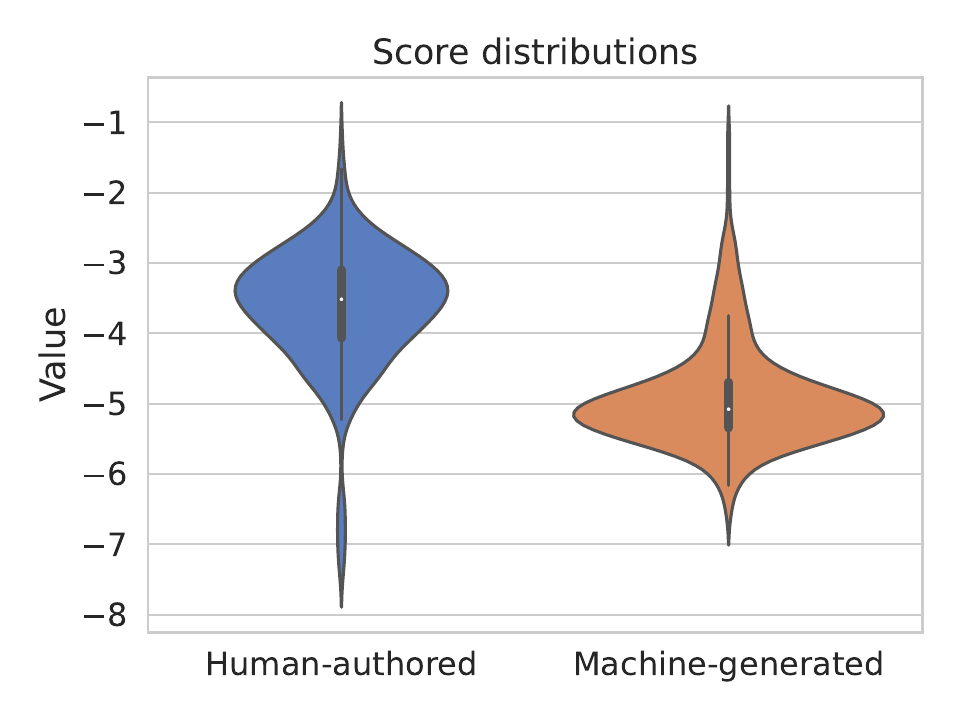}
}
\subfigure[Log Rank of XSum.]{
\includegraphics[width=.23\textwidth]{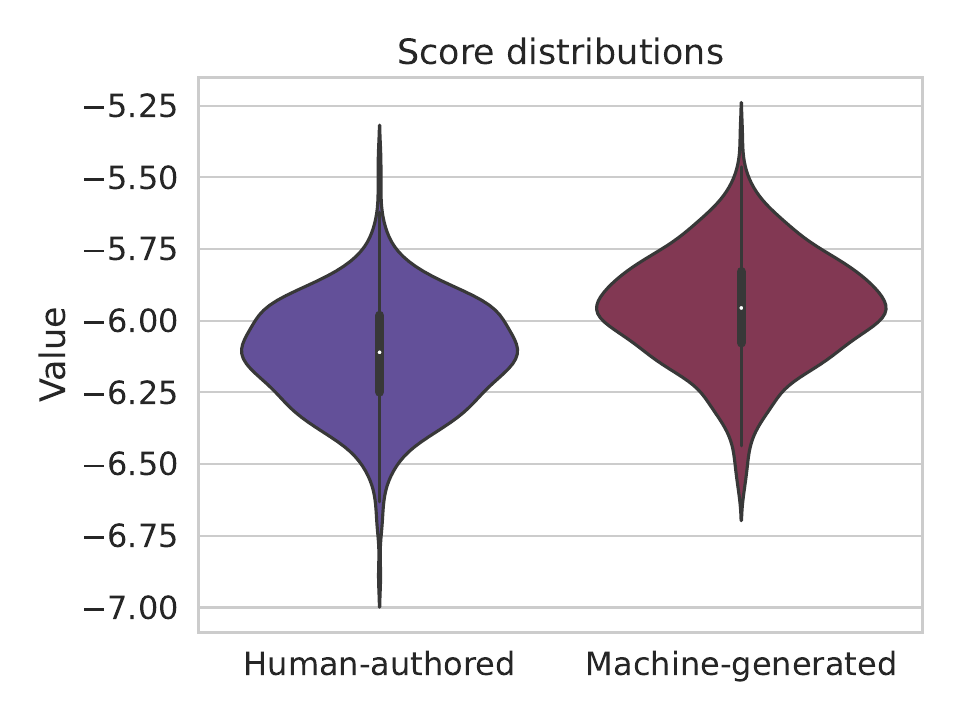} 
}
\subfigure[Log Rank of SQuAD.]{
\includegraphics[width=.23\textwidth]{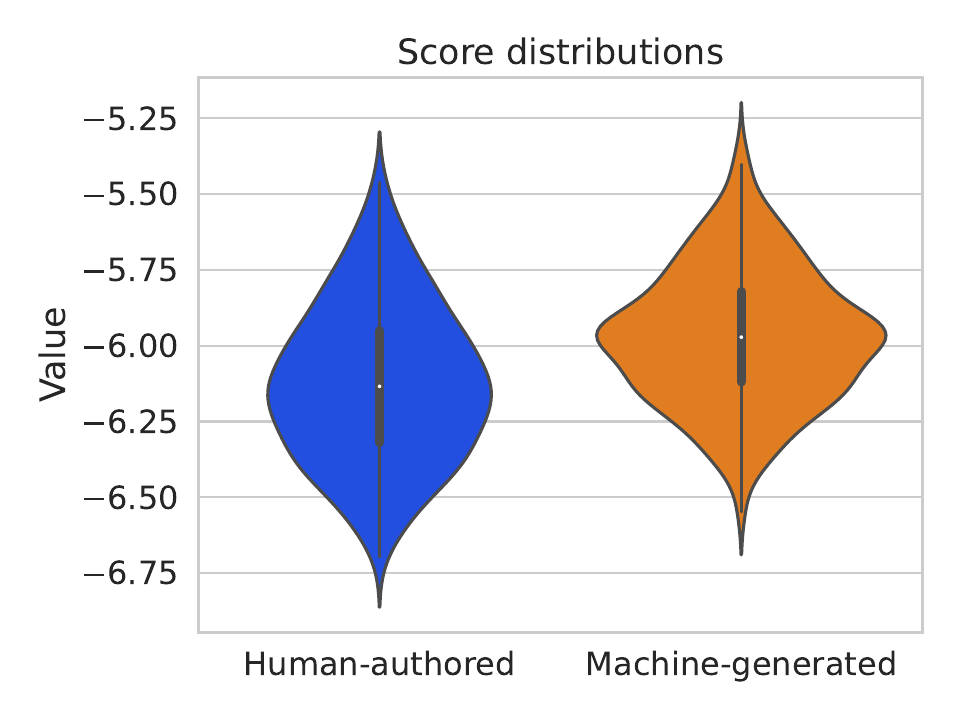} 
}
\subfigure[Log Rank of Writing.]{
\includegraphics[width=.23\textwidth]{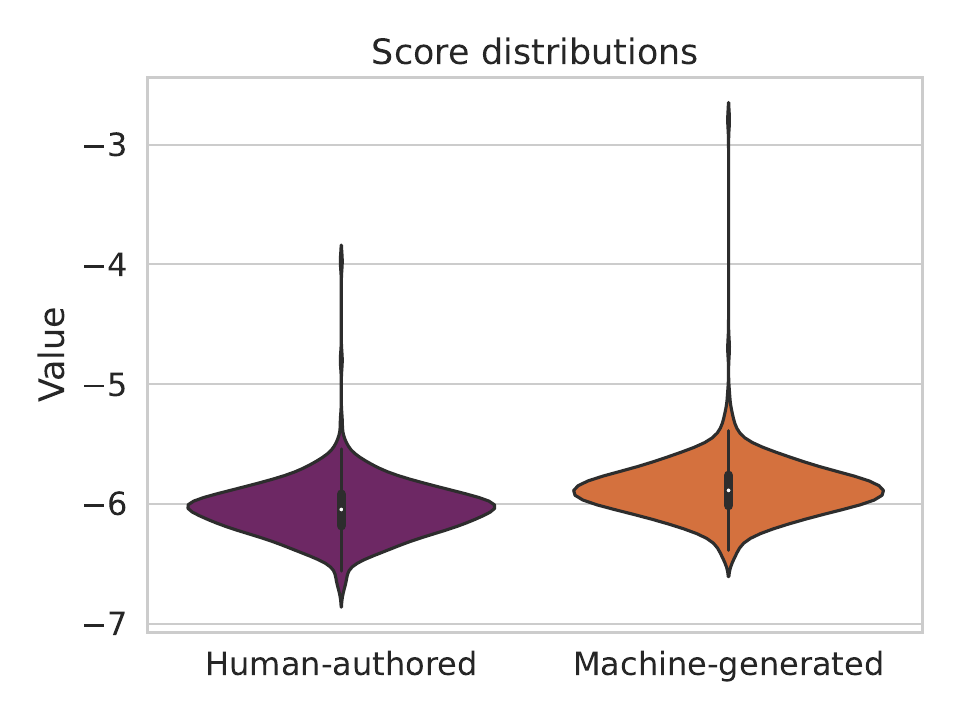} 
}
\end{minipage}
\caption{
\textbf{Score distributions} of log probabilities (a,b,c,d)  rank (e,f,g,h), and log rank (i,j,k,l) considering different topics.
}
\label{fig:app_socres}
\vspace{-0.2cm}
\end{figure*}

\end{document}